\documentclass[letterpaper, 10 pt, conference]{ieeeconf}  

\IEEEoverridecommandlockouts                              

\usepackage{amsmath,amsfonts,bm}

\def\1{\bm{1}}

\DeclareMathAlphabet{\mathsfit}{\encodingdefault}{\sfdefault}{m}{sl}
\SetMathAlphabet{\mathsfit}{bold}{\encodingdefault}{\sfdefault}{bx}{n}

\def\gA{{\mathcal{A}}}

\def\gR{{\mathcal{R}}}
\def\gS{{\mathcal{S}}}

\usepackage{algorithm}
\usepackage{algpseudocode}
\usepackage{graphics} 
\usepackage{epsfig} 
\usepackage{amsmath} 
\usepackage{amssymb}  
\usepackage{booktabs}
\usepackage{balance}
\usepackage[hyphens]{xurl}
\usepackage[hidelinks]{hyperref}

\title{\LARGE \bf
Adaptive Linear Path Model-Based Diffusion
}

\author{Yutaka Shimizu and Masayoshi Tomizuka
\thanks{Yutaka Shimizu and Masayoshi Tomizuka are with the Department of Mechanical Engineering, University of California, Berkeley, CA 94720, USA        {\tt\small \{purewater0901, tomizuka\}@berkeley.edu}}
}

\begin{document}

\maketitle
\thispagestyle{empty}
\pagestyle{empty}

\begin{abstract}
The interest in combining model-based control approaches with diffusion models has been growing. Although we have seen many impressive robotic control results in difficult tasks, the performance of diffusion models is highly sensitive to the choice of scheduling parameters, making parameter tuning one of the most critical challenges.
We introduce Linear Path Model-Based Diffusion (LP-MBD), which replaces the variance-preserving schedule with a flow-matching–inspired linear probability path. This yields a geometrically interpretable and decoupled parameterization that reduces tuning complexity and provides a stable foundation for adaptation. Building on this, we propose Adaptive LP-MBD (ALP-MBD), which leverages reinforcement learning to adjust diffusion steps and noise levels according to task complexity and environmental conditions. Across numerical studies, Brax benchmarks, and mobile-robot trajectory tracking, LP-MBD simplifies scheduling while maintaining strong performance, and ALP-MBD further improves robustness, adaptability, and real-time efficiency.
Our code is available through anonymous repository \url{https://anonymous.4open.science/r/adaptive_linear_path_model_based_diffusion-C58C}
\end{abstract}

\section{INTRODUCTION}

Diffusion models have achieved remarkable success in various real-world applications, particularly in image and video generation. Recently, their use has expanded beyond visual domains, with a growing body of work exploring applications in robotic control. Several approaches \cite{pan2024modelbased} \cite{dial-mpc} have been proposed that integrate diffusion models with model-based control to solve complex trajectory optimization problems. These approaches have close relationships with sampling-based optimization methods, and they have been shown to be effective in addressing problems that have nonlinear, non-smooth dynamics and non-convex objectives and constraints.

\begin{figure}[htbp]
  \begin{minipage}[b]{0.46\linewidth}
    \centering
    \includegraphics[width=\linewidth]{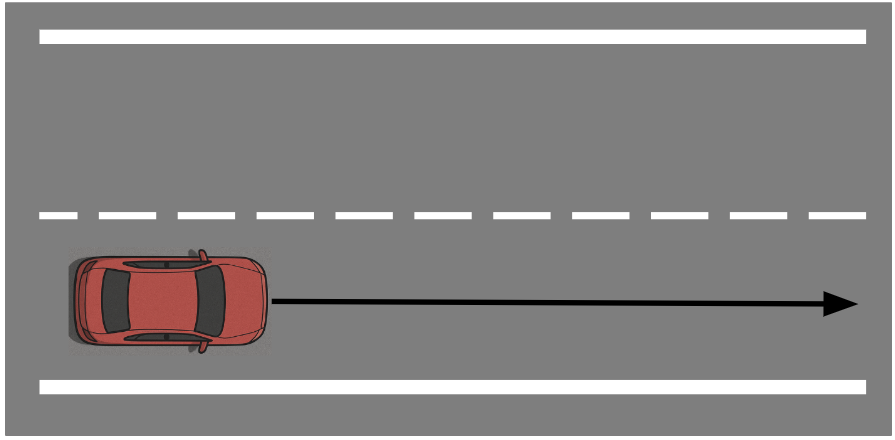}
  \end{minipage}
  \hspace{0.03\linewidth}
  \begin{minipage}[b]{0.46\linewidth}
    \centering
    \includegraphics[width=\linewidth]{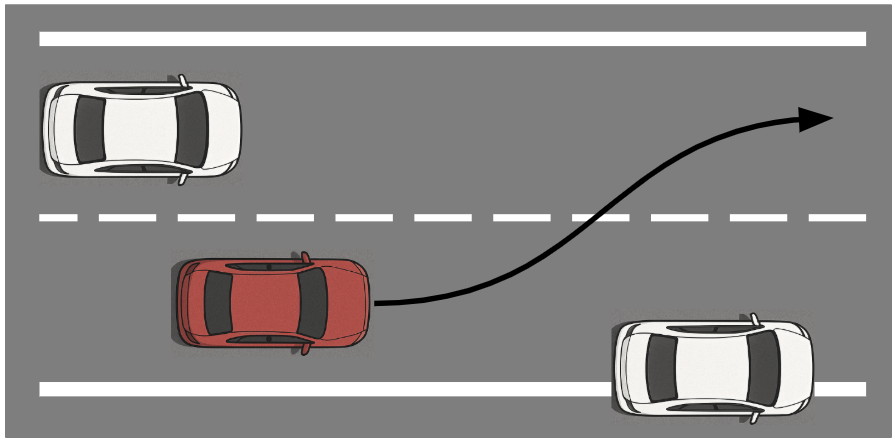}
  \end{minipage}
  \caption{The left figure illustrates a simple scenario in which the red ego vehicle drives in an obstacle-free environment. The right figure depicts a more complex case, where the red ego vehicle attempts to overtake a white vehicle while simultaneously avoiding an approaching car from behind. In the latter scenario, additional safety constraints are imposed, typically requiring more diffusion steps to obtain an optimal trajectory.}
  \label{fig:scenario-exxample}
\end{figure}

However, the performance of model-based diffusion critically depends on the design of the noise schedule. 
In diffusion-based approaches, the sampling process consists of a sequence of diffusion steps, where noise is incrementally injected into trajectories and progressively reduced to produce feasible and improved samples.
The scale of the injected noise at each step plays an important role in determining performance, and even in the case of a simple linear parameterization with a variance-preserving schedule \cite{pan2024modelbased}, we need to tune several parameters to get the optimal solution. These parameters are intricately coupled in shaping the overall noise profile, making it difficult to find optimal values.
Moreover, the optimal parameters often vary with the system state and the complexity of the problem, which frequently leads to overly conservative parameter choices. Even within the same task, the values of the optimal parameters can differ substantially across different conditions. We illustrate this idea by using a self-driving car example shown in Fig.~\ref{fig:scenario-exxample}.

In this paper, we first introduce Linear Path Model-Based Diffusion (LP-MBD), a variant of the original Model-Based Diffusion (MBD) framework that replaces the variance-preserving schedule with a linear probability path inspired by Flow Matching \cite{lipman2023flowmathcing}\cite{lipman2024flowmatchingguidecode}.
LP-MBD offers several advantages over the original MBD. 
The linear probability path provides a simple, geometrically interpretable interpolation between prior and target distributions; under common Gaussian assumptions, it can align with optimal transport, offering clear theoretical grounding.
More importantly, unlike the variance-preserving schedule, where multiple parameters are intricately coupled to determine the noise magnitude, the linear path formulation eliminates such dependencies. As a result, LP-MBD requires fewer hyperparameters, making the tuning process more straightforward and interpretable.

Building upon this foundation, we further propose Adaptive Linear Path Model-Based Diffusion (ALP-MBD), which extends LP-MBD with a reinforcement learning–based module for dynamic parameter adjustment. ALP-MBD adapts scheduling parameters according to task complexity and environmental conditions. For instance, it can increase the number of diffusion steps in challenging scenarios or enlarge the noise level to promote broader exploration, thereby enabling more flexible and effective control.
In contrast, under simple or well-structured conditions, it can reduce both the noise magnitude and the diffusion steps, leading to more efficient sampling without sacrificing solution quality.

We extensively evaluate LP-MBD and ALP-MBD across diverse environments and settings to demonstrate their efficiency and performance. Our experiments include numerical studies, Mujoco-based benchmarks implemented in Brax, and a mobile robot trajectory-tracking task. 

This paper has three main contributions:
\begin{enumerate}
    \item \textbf{Linear Path Model-Based Diffusion (LP-MBD):} We introduce a flow-matching–inspired linear path scheduler that yields a geometrically interpretable and decoupled parameterization. This reduces the burden of tuning and provides a stable foundation for adaptive extensions.
    
    \item \textbf{Adaptive Linear Path MBD (ALP-MBD):} Building on LP-MBD, we develop an adaptive scheduler that leverages reinforcement learning to adjust diffusion steps and noise levels based on the environment state, enhancing both robustness and efficiency.
    
    \item \textbf{Comprehensive evaluation:} We validate LP-MBD and ALP-MBD through numerical studies, Brax benchmarks, and mobile-robot trajectory tracking, demonstrating improved sample efficiency, control quality, and adaptability across diverse settings.
\end{enumerate}

\section{Related Work}
Trajectory optimization is a fundamental component of safe and precise robot control. There are two principal approaches: (i) solving the underlying optimization problem directly with gradient-based methods \cite{Hargraves1987DirectTO, Altro, Kim2007AnIM, Shi2022Neural}, and (ii) employing sampling-based approaches \cite{DynamicWindow, Howard2008State} to search for optimal trajectories. This paper focuses on the latter, as it can better accommodate complex dynamics and objective functions.
 MPPI \cite{mppi1, mppi2} is among the most widely used sampling algorithms.
It can be viewed as a soft-weighted variant of the Cross-Entropy Method (CEM) \cite{CEM}, where samples are weighted smoothly rather than selected by a hard elite set.
CEM itself is closely related to CMA-ES \cite{cma-es}, but CMA-ES further refines the update by adapting the covariance and step size in a principled way, making it more robust for problems with strong correlations or anisotropy.

Generative models provide powerful tools for capturing complex distributions. Diffusion models \cite{DDPM, DDIM, sdxl} and Flow Matching \cite{lipman2023flowmathcing, lipman2024flowmatchingguidecode} have gained significant attention due to their ability to represent intricate distributions and have been applied to planning problems \cite{diffusion_policy1, diffusion_policy2, janner2022diffuser, safe_flow}. Recent works such as \cite{Berner2022AnOC, kurtz2025generative} explore the link between sampling-based methods and diffusion models, while \cite{dial-mpc} highlights the connection between diffusion models and MPPI.

Reinforcement Learning (RL) \cite{SuttonRL} has been applied across diverse domains, including model-based planning in combination with sampling-based planners \cite{TDMPC, TDMPC2, BSMPC}. These studies demonstrate that integrating RL with sampling methods achieves superior performance compared to state-of-the-art model-free approaches \cite{CURL, sac, DPG, ddpg, td3}. Beyond planning, RL has also been used to fine-tune diffusion models \cite{ueharaICML2024, clark2024directly, DPOK}. For instance, \cite{wagenmaker2025steering} employs RL to optimize the initial noise in diffusion models, enabling mode selection within complex distributions. In this paper, we leverage RL to adaptively determine scheduling parameters conditioned on the environment state.

\section{Linear Path Model-Based Diffusion}
In this section, we first review Model-Based Diffusion (MBD) \cite{pan2024modelbased} and discuss why the choice of diffusion schedule can become a key practical bottleneck. We subsequently present Linear Path MBD (LP-MBD), which incorporates the linear probability path from Flow Matching. By constructing an optimal transport interpolation between the prior and target distributions, this formulation not only simplifies hyperparameter tuning but also provides a more stable and geometrically interpretable trajectory generation process.

\subsection{Variance Preserving Model-Based Diffusion}
Trajectory optimization (TO) is a fundamental tool for steering a robot toward a desired goal. We consider a finite-horizon problem with states $x_t \in \mathbb{R}^{n_x}$ and controls $u_t \in \mathbb{R}^{n_u}$:
\begin{subequations} \label{eq:trajectory-optimization}
\begin{align}
& \min_{x_{1:T},\,u_{1:T}}~ J(x_{1:T};u_{1:T}) \\
& \text{s.t.}\quad x_{t+1}=f_t(x_t,u_t), \\
& \qquad\qquad g_t(x_t,u_t)\le 0,\quad t=0,\dots,T-1,
\end{align}
\end{subequations}
where $J$ is a user-specified cost, $f_t$ denotes the system dynamics, and $g_t$ is constraints. Let $Y=[x_{1:T};u_{1:T}]$ denote the decision variables. Following \cite{pan2024modelbased}, we recast TO as a sampling problem from the following target probability distribution
\begin{align} \label{eq:probability-density}
p_0(Y)\;\propto\; p_d(Y)\,p_J(Y)\,p_g(Y),
\end{align}
where $p_d$ enforces dynamical feasibility, $p_g$ encodes constraints, and $p_J(Y)\propto \exp\!\big(-J(Y)/\lambda\big)$ biases toward low cost with temperature $\lambda>0$.

MBD \cite{pan2024modelbased} samples decision variables using a diffusion process. A standard forward process gradually perturbs an initial distribution $p_0$ toward an isotropic Gaussian $p_N$ with noise governed by a schedule $\{\alpha_i\}_{i=1}^N$:
\begin{gather}
\label{eq:forward-diffusion-process}
Y^{(i)} = c_{i, 0} \, Y^{(0)} + c_{i, 1}\, \varepsilon, \\
\label{eq:vp-setting}
c_{i,0} = \sqrt{\bar\alpha_i}, \quad c_{i,1} = \sqrt{(1-\bar\alpha_i)}, \quad \bar\alpha_i=\prod_{k=1}^{i}\alpha_k,
\end{gather}
where $\varepsilon \sim \mathcal{N}(0, I)$. Here, $Y^{(0)}$ denotes a sample drawn from the initial distribution $p_0$, which corresponds to the original decision variables of the optimization problem before any perturbation is applied. 
Note that the coefficients $c_{i,0}$ and $c_{i,1}$ define the noise schedule of the forward process and thus critically influence the performance of the diffusion model.

From Eq.~\eqref{eq:forward-diffusion-process} and Eq.~\eqref{eq:vp-setting}, if the variance of the initial distribution is standardized such that $\mathrm{Var}[Y^{(0)}]=1$, the variance of $Y^{(i)}$ remains equal to one for all $i$. This invariance arises because the coefficients $c_{i,0}$ and $c_{i,1}$ are chosen to satisfy $c_{i,0}^2 + c_{i,1}^2 = 1$, thereby preserving the overall variance throughout the forward diffusion process. For this reason, the schedule $\{\alpha_i\}$ is referred to as a variance-preserving (VP) noise schedule. 
In practice, MBD uses a simple linear VP schedule in which $\beta_i$ is interpolated linearly over $i=1,\dots,N$ with $\beta_0=1.0\times 10^{-4}$ and $\beta_1=1.0\times 10^{-2}$, and $\alpha_i=1-\beta_i$. We denote this specific instantiation of MBD as VP-MBD, in order to distinguish it from the proposed approach.

Unlike model-free diffusion planners that learn the score from data, MBD exploits known objectives and dynamics to estimate the score $\nabla_{Y^{(i)}}\log p_i(Y^{(i)})$ and performs a Monte-Carlo score-ascent–type denoising step:
\begin{equation} \label{eq:mbd-sampling}
\begin{split}
Y^{(i-1)} &= \frac{c_{i-1, 0}}{c_{i, 0}} \Big(Y^{(i)} + c_{i, 1}^2\,\nabla_{Y^{(i)}}\log p_i\!\big(Y^{(i)}\big)\Big) \\
&= \frac{1}{\sqrt{\alpha_i}} \Big(Y^{(i)} + (1-\bar\alpha_i)\,\nabla_{Y^{(i)}}\log p_i\!\big(Y^{(i)}\big)\Big)
\end{split}
\end{equation}
The score can be written (via Bayes’ rule and the forward kernel) as an expectation over ``clean'' trajectories $Y^{(0)}$ sampled from a Gaussian proposal and reweighted by the model-based target:
\begin{equation} \label{eq:mbd-score-calculation}
\begin{split}
& \nabla_{Y^{(i)}}\log  p_i\!\big(Y^{(i)}\big) ~\approx~  \\ 
& \; -\frac{Y^{(i)}}{1-\bar\alpha_i} +\frac{\sqrt{\bar\alpha_i}}{1-\bar\alpha_i}\ \underbrace{\frac{\sum_{Y^{(0)}\in\mathcal{Y}^{(i)}_{\text{VP-MBD}}} Y^{(0)}\, w(Y^{(0)})}{\sum_{Y^{(0)}\in\mathcal{Y}^{(i)}_{\text{VP-MBD}}} w(Y^{(0)})}}_{\displaystyle \text{ (importance-weighted average)}}
\end{split}
\end{equation}
where $w(Y)=p_J(Y)\,p_g(Y)$ and $\mathcal{Y}^{(i)}_{\text{VP-MBD}}$ follows a Gaussian distribution:
\begin{equation} \label{eq:mbd-gaussian-proposal}
\mathcal{Y}^{(i)}_{\text{VP-MBD}} \sim \mathcal{N}\Big(\frac{Y^{(i)}}{\sqrt{\bar\alpha_i}}, \frac{1-\bar\alpha_i}{\bar\alpha_i}\Big) 
\end{equation}
For TO, candidate $Y^{(0)}=[x_{1:T};u_{1:T}]$ are made dynamically feasible by rolling out $x_{t+1}=f_t(x_t,u_t)$ (shooting), and then scored via $w(Y)$. Substituting Eq.~\eqref{eq:mbd-score-calculation} into Eq.~\eqref{eq:mbd-sampling}, we get 
\begin{equation} \label{eq:mbd-update-equation}
    Y^{(i-1)} = \sqrt{\bar\alpha_{i-1}} \frac{\sum_{Y^{(0)}\in\mathcal{Y}^{(i)}_{\text{VP-MBD}}} Y^{(0)}\, w(Y^{(0)})}{\sum_{Y^{(0)}\in\mathcal{Y}^{(i)}_{\text{VP-MBD}}} w(Y^{(0)})}
\end{equation}

Although VP-MBD attains strong performance, tuning the noise scheduling parameters remains challenging. In the variance-preserving (VP) formulation with simple linear scheduling, the schedule is specified by the triplet $(\beta_0,\beta_1,T)$, which jointly determine $\{\alpha_i\}$ and the cumulative noise levels $\{\bar\alpha_i\}$. In particular, the maximum noise level is not controlled solely by the endpoints $\beta_0$ and $\beta_1$, but is also influenced by the total number of diffusion steps $T$. As a result, these parameters interact in a nontrivial manner to shape the effective noise scale at each step. This interdependence complicates the tuning process, as the impact of each parameter on the exploration–refinement trade-off is highly indirect and task-dependent, often requiring extensive trial-and-error to obtain satisfactory performance.

\subsection{Linear Path Model-Based Diffusion}
Motivated by flow matching, we adopt a \emph{linear probability path} between a clean trajectory $Y^{(0)}$ and a standard Gaussian $\varepsilon \!\sim\!\mathcal{N}(0,I)$:
\begin{equation} \label{eq:linear-continuous-path}
Y_t \;=\; (1-t)\,Y^{(0)} \;+\; t\,\varepsilon,\qquad t\in[0,1].
\end{equation}
Discretizing $t$ on a uniform grid $t_i=\tfrac{i}{N}$, $i=0,\dots,N$, yields a schedule that increases the noise level linearly in ``time.''
\begin{equation} \label{eq:linear-discretized-path}
Y^{(i)} \;=\; (1-t_i)\,Y^{(0)} \;+\; t_i\,\varepsilon
\end{equation}
This corresponds to Eq.~\eqref{eq:forward-diffusion-process} when $c_{i,0} = 1-t_i$ and $c_{i,1} = t_i$.
At the endpoints, $i=0$ ($t_i=0$) gives $Y^{(i)}=Y^{(0)}$ (clean trajectory), while $i=N$ ($t_i=1$) yields $Y^{(N)}=\varepsilon$ (standard Gaussian noise).
Thus, the linear probability path recovers the clean sample at one endpoint and pure noise at the other, providing a clear and interpretable interpolation between the two distributions.

We use the same Monte-Carlo score-ascent-type denoising step:
\begin{equation} \label{eq:linear-score-ascent}
Y^{(i-1)} = \frac{1-t_{i-1}}{1-t_i} \Big(Y^{(i)} + t_i^2\,\nabla_{Y^{(i)}}\log p_i\!\big(Y^{(i)}\big)\Big)
\end{equation}
where the score function is
\begin{equation} \label{eq:linear-score-calculation}
\begin{split}
& \nabla_{Y^{(i)}}\log  p_i\!\big(Y^{(i)}\big) ~\approx~  \\ 
& \; -\frac{Y^{(i)}}{t_i^2} +\frac{1-t_i}{t_i^2}\ \underbrace{\frac{\sum_{Y^{(0)}\in\mathcal{Y}^{(i)}_{\text{LP-MBD}}} Y^{(0)}\, w(Y^{(0)})}{\sum_{Y^{(0)}\in\mathcal{Y}^{(i)}_{\text{LP-MBD}}} w(Y^{(0)})}}_{\displaystyle \text{ (importance-weighted average)}} 
\end{split}
\end{equation}
Similar to VP-MBD, we sample $Y^{(0)}$ from the following Gaussian distribution
\begin{equation} \label{eq:lp-mbd-gaussian-proposal}
    \mathcal{Y}^{(i)}_{\text{LP-MBD}} \sim \mathcal{N}\Big(\frac{Y^{(i)}}{1-t_i}, \frac{t_i^2}{(1-t_i)^2}\Big)
\end{equation}
By substituting Eq.~\eqref{eq:linear-score-calculation} into Eq.~\eqref{eq:linear-score-ascent}, we get
\begin{equation} \label{eq:linear-mbd-update-equation}
    Y^{(i-1)} = (1-t_{i-1}) \frac{\sum_{Y^{(0)}\in\mathcal{Y}^{(i)}_{\text{LP-MBD}}} Y^{(0)}\, w(Y^{(0)})}{\sum_{Y^{(0)}\in\mathcal{Y}^{(i)}_{\text{LP-MBD}}} w(Y^{(0)})}
\end{equation}

Although LP-MBD uses an intuitive noise schedule, the Gaussian proposal in Eq.~\eqref{eq:lp-mbd-gaussian-proposal} has a standard deviation
$\sigma_i=\tfrac{t_i}{1-t_i}$, which diverges as $t_i\to 1$.
This implies that the initial backward denoising step samples the entire trajectory space, which is theoretically valid but practically unnecessary.
In trajectory optimization, control inputs are typically bounded by system constraints, so sampling from an unbounded domain is inefficient. Constraining the noise schedule to natural limits improves efficiency without loss of correctness, motivating a bounded scheduling strategy.  
Instead of extending the interpolation to $t_i=1.0$, we truncate the schedule at a maximum value $t_{\max}<1$, ensuring that the variance of $\mathcal{Y}^{(i)}_{\text{LP-MBD}}$ remains finite: 
\begin{equation}
    t_i \in [0,\,t_{\max}], \qquad t_{\max}<1.
\end{equation}
When $t_i = t_{\max}$, the standard deviation of the Gaussian distribution $\mathcal{Y}^{(i)}_{\text{LP-MBD}}$ reaches its maximum value.
\begin{equation} \label{eq:max-standard-deviation}
    \sigma_{\max} \;=\; \frac{t_{\max}}{1-t_{\max}}
  \quad \left( \text{equivalently, } t_{\max}=\frac{\sigma_{\max}}{1+\sigma_{\max}} \right)
\end{equation}
In practice, $\sigma_{\max}$ can be determined from the admissible range of control inputs, which provides a direct and interpretable way to set the exploration limit. Therefore, $t_{\max}$ can be computed from Eq.~\eqref{eq:max-standard-deviation}.

\subsection{Differences between VP-MBD and LP-MBD}
The primary distinction between VP-MBD and LP-MBD lies in their parameterization and the implications for tuning and adaptation. In the VP-MBD formulation, the noise schedule is governed by three parameters $(\beta_0, \beta_1, T)$, where the parameters are tightly coupled: modifying the diffusion horizon $T$ not only changes the discretization but also increases or decreases the maximum noise variance. This interdependence complicates the tuning process and makes it difficult to isolate the effect of each parameter on exploration and refinement.

In contrast, LP-MBD is characterized by only two parameters: the maximum noise scale $\sigma_{\max}$ and the number of diffusion steps $T$. These parameters are \emph{decoupled} and possess clear geometric interpretations: $\sigma_{\max}$ directly sets the maximum variance of the Gaussian proposal, while $T$ controls the resolution of the interpolation. Importantly, adjusting $T$ does not affect the extrema of the noise scale. This structural simplicity makes the noise schedule not only easier to tune but also more geometrically interpretable, providing an intuitive understanding of the interpolation between clean trajectories and noise.
Beyond simplifying tuning, this decoupling also enables the adaptive extension in the next section. With two independent parameters, the parameter adaptation approach can efficiently optimize them, making LP-MBD a solid foundation for ALP-MBD.

\section{Adaptive Linear Path Model-Based Diffusion}
In VP-MBD, the diffusion steps and noise scheduling parameters are fixed once selected and remain unchanged during execution. In practice, however, the difficulty of trajectory optimization can vary substantially even within the same task. For instance, navigation in open space may require a few diffusion steps, while obstacle-rich scenarios demand more refinement and broader exploration. This motivates adapting the diffusion process online to balance robustness and efficiency.

The decoupled and geometrically interpretable parameterization of LP-MBD provides a natural foundation for such adaptation. Unlike VP-MBD, where the parameters $(\beta_0, \beta_1, T)$ are tightly coupled and changes in $T$ also influence the effective noise scale, LP-MBD separates the maximum variance $\sigma_{\max}$ from the diffusion horizon $T$. This decoupling ensures that adjusting $T$ only refines the discretization without unintentionally changing the exploration range. As a result, the training process of learning optimal parameters of LP-MBD is more stable and simple. In contrast, applying RL to VP-MBD would face instability, since parameter updates may produce unpredictable changes in the underlying noise profile. Thus, LP-MBD is inherently more compatible with adaptive parameter learning, enabling the design of Adaptive LP-MBD (ALP-MBD).

\subsection{Formulation as a Reinforcement Learning Problem}
Reinforcement learning (RL) aims to optimize decision-making policies in environments modeled as a Markov Decision Process (MDP) $(\gS, \gA, \mathcal{P}, \gR, \rho_0, \gamma)$. 
Here, $\gS$ denotes the state space, $\gA$ the action space, $\mathcal{P}$ the transition dynamics, $\gR$ the reward function, $\rho_0$ the initial state distribution, and $\gamma$ the discount factor. 
At each time step, the agent observes $s \in \gS$, selects an action $a \in \gA$, transitions to $s' \sim \mathcal{P}(\cdot|s,a)$, and receives a reward $r(s,a)$.

In our formulation, the state space $\gS$ corresponds to the environment state as in standard RL. 
However, unlike conventional RL where actions directly control the system, the action space $\gA$ consists of the noise scheduling parameters $T$ and $\sigma_{\max}$. 
The actual control signal $u$ applied to the system is subsequently generated by LP-MBD with these estimated parameters. 
Accordingly, the adaptive noise scheduler serves as a policy that, at each step $t$, outputs
\begin{equation} \label{eq:diffusion-step-sigma-max-distribution}
\begin{split}
    (T_t,\;\sigma_{\max,t}) &\sim \pi_\phi(\cdot \mid s_t), \\
    \pi_\phi(T,\sigma_{\max}\mid s) &= \pi_T(T \mid s)\, \pi_\sigma(\sigma_{\max}\mid s),
\end{split}
\end{equation}
where $\phi$ denotes the policy parameters, $\pi_T$ is a categorical distribution over $\mathcal{T}=\{T_{\min}, \ldots, T_{\max}\}$, and $\pi_\sigma$ is a Gaussian distribution. 
The action at time step $t$ is therefore defined as $a_t=(T_t, \sigma_{\max,t})$.

\begin{figure}[ht]
  \centering
  \includegraphics[scale=0.33]{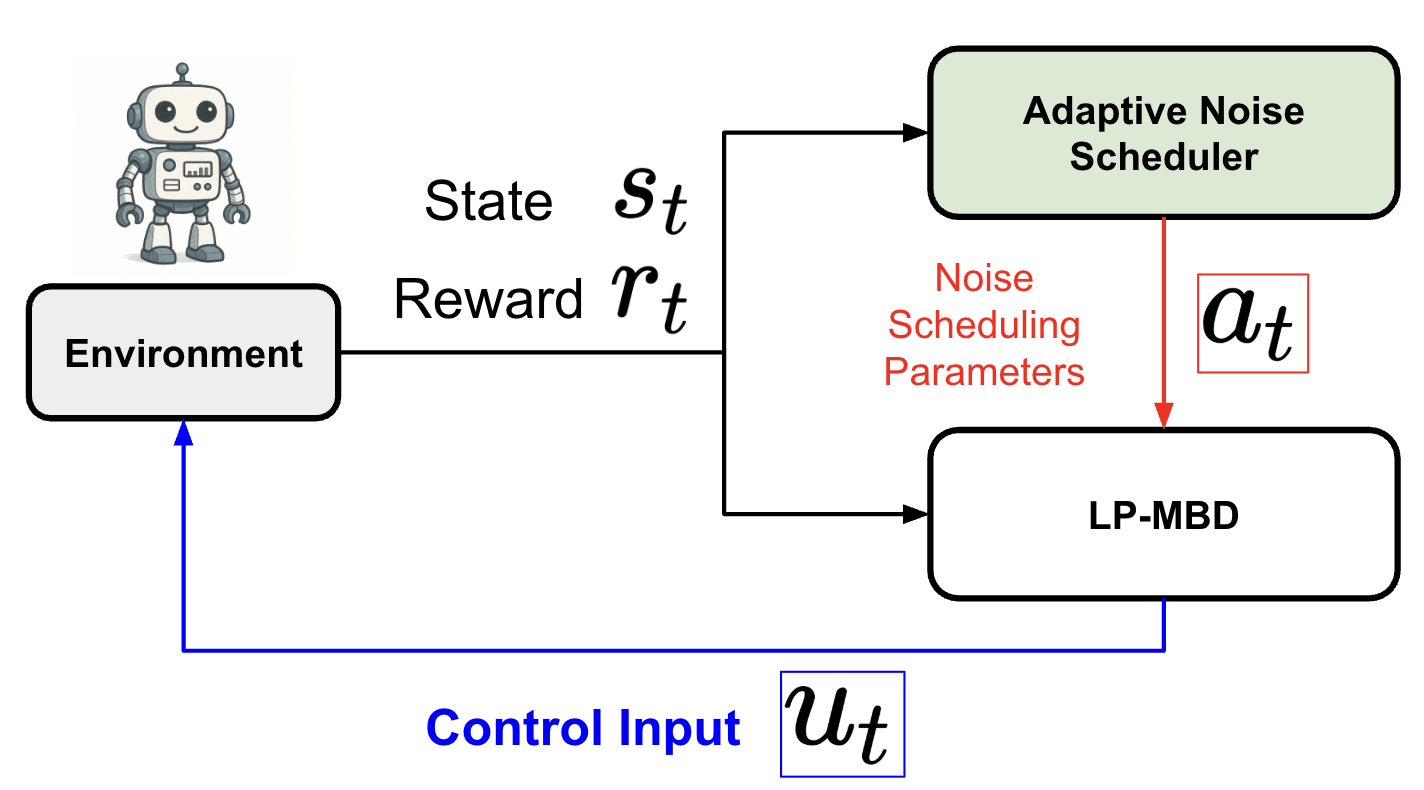}
  \caption{Overview of ALP-MBD. 
The environment provides the current state $s_t$ and reward $r_t$ to the adaptive noise scheduler, which outputs the noise scheduling parameters $a_t = (T_t, \sigma_{\max,t})$. 
These parameters are fed into LP-MBD to generate the control input $u_t$, which is applied to the environment.}
  \label{fig:adaptive-noise-scheduler}
\end{figure}

To promote efficiency, we augment the reward with a penalty on large diffusion steps:
\begin{equation} \label{eq:augmented-reward}
\tilde{r}_t(s_t, a_t, u_t) = r(s_t,u_t) - w_T \frac{T_t}{T_{\max}},
\end{equation}
where $r(s_t, u_t)$ is the original environmental reward and $w_T>0$ balances task performance and computational cost. 
Denoting LP-MBD as $\pi_{\text{LP-MBD}}(\cdot \mid s_t, a_t)$, the RL objective is
\begin{equation}
\max_{\phi} J(\phi) = 
\mathbb{E}_{\substack{s_t \sim \mathcal{B},\, a_t \sim \pi_{\phi}(\cdot \mid s_t), \\ u_t \sim \pi_{\text{LP-MBD}}(\cdot \mid s_t, a_t)}} 
\left[ \sum_{t=0}^\infty \gamma^t \tilde r_t \right],
\end{equation}
with discount factor $\gamma \in (0,1)$ and replay buffer $\mathcal{B}$.
Fig.~\ref{fig:adaptive-noise-scheduler} shows the architecture of the proposed adaptive noise scheduling system.

\subsection{Implementation Details}
To train the adaptive linear path scheduler, we employ the Proximal Policy Optimization (PPO) \cite{PPO} algorithm. 
PPO is chosen for its stability in on-policy training and its widespread adoption in both control and discrete tasks. 
Nevertheless, the proposed framework is agnostic to the choice of reinforcement learning algorithm, and other approaches (e.g., DDPG\cite{ddpg}, SAC\cite{sac}\cite{sac2}\cite{sac3}, or TD3\cite{td3}) can also be applied without loss of generality.

We summarize the training procedure in Algorithm~\ref{alg:adaptive_scheduler}. 
At each iteration, the adaptive noise scheduler samples scheduling parameters $(T_t, \sigma_{\max,t})$ according to the current state, which are then used by the LP-MBD to generate the control input $u_t$. 
The environment returns the next state $s_{t+1}$ and reward $r_t$, which is reshaped into $\tilde{r}_t$ to penalize large diffusion steps. 
The PPO algorithm then updates the policy parameters $\phi$ and value function parameters $\psi$ using collected trajectories from a rollout buffer.

\begin{algorithm}[ht]
\caption{Training ALP-MBD with PPO}
\label{alg:adaptive_scheduler}
\begin{algorithmic}[1]
\State Initialize policy parameters $\phi$, value function parameters $\psi$
\For{each iteration}
    \For{each environment step $t$}
        \State Observe state $s_t$
        \State Sample scheduling parameters $a_t=(T_t,\sigma_{\max,t}) \sim \pi_\phi(\cdot|s_t)$
        \State Generate control input $u_t \sim \pi_{\text{LP-MBD}}(\cdot|s_t,a_t)$
        \State Apply $u_t$ to environment, receive $(s_{t+1}, r_t)$
        \State Compute modified reward Eq.~\eqref{eq:augmented-reward}
        \State Store $(s_t, a_t, u_t, \tilde r_t, s_{t+1})$ in buffer $\mathcal{B}$
    \EndFor
    \State Use PPO update to optimize $\phi, \psi$ with data from $\mathcal{B}$
    \State Clear buffer $\mathcal{B}$
\EndFor
\end{algorithmic}
\end{algorithm}

\section{Experiments}
In this section, we conduct a series of experiments to evaluate the proposed LP-MBD and ALP-MBD. 
These experiments are designed to validate the following aspects: (i) the behavioral and performance differences between VP-MBD and LP-MBD; (ii) the advantages of LP-MBD in reducing the tuning burden; and (iii) the ability of ALP-MBD to adjust key parameters—primarily the diffusion horizon $T$ and the maximum noise standard deviation $\sigma_{\max}$—as functions of the observed state, and the extent to which these adaptations improve efficiency.
All experiments were conducted on a single NVIDIA RTX 4090 GPU.
Throughout this experiment, we use $\beta_0=0.0001$ and $\beta_1=0.01$ for VP-MBD and $\sigma_{\max} = 1.8$ for LP-MBD.

\subsection{Numerical Experiments}
We first perform simple numerical experiments to evaluate LP-MBD and ALP-MBD and compare with VP-MBD, illustrating their relative advantages in a controlled setting. 
Our numerical studies (i) show why VP-MBD parameter tuning is harder than LP-MBD and (ii) illustrate how LP-MBD’s simple and independent parameter sets better support adaptive scheduling.

\begin{figure}[htbp] 
    \centering
    \includegraphics[scale=0.17]{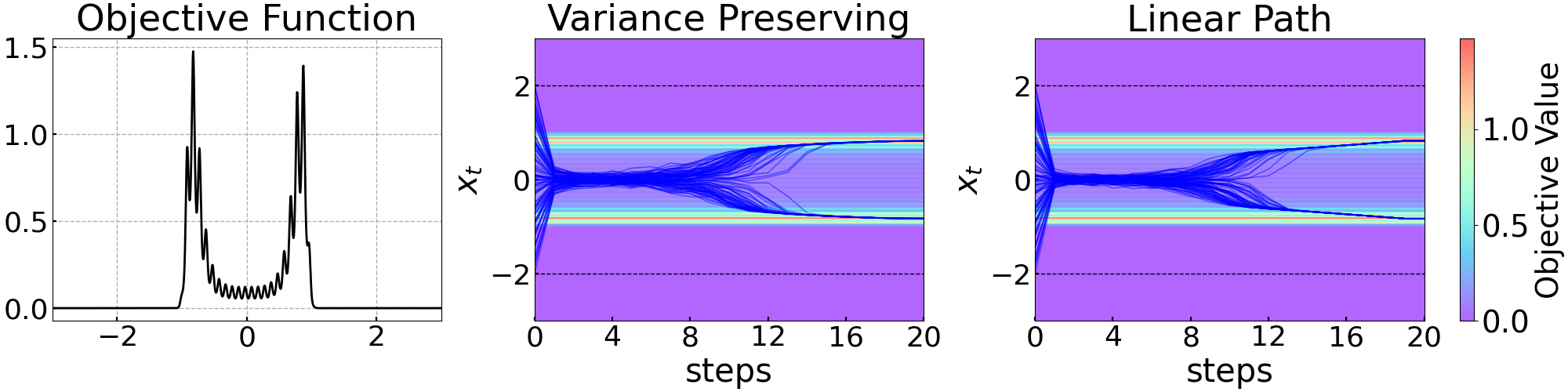}
    \includegraphics[scale=0.17]{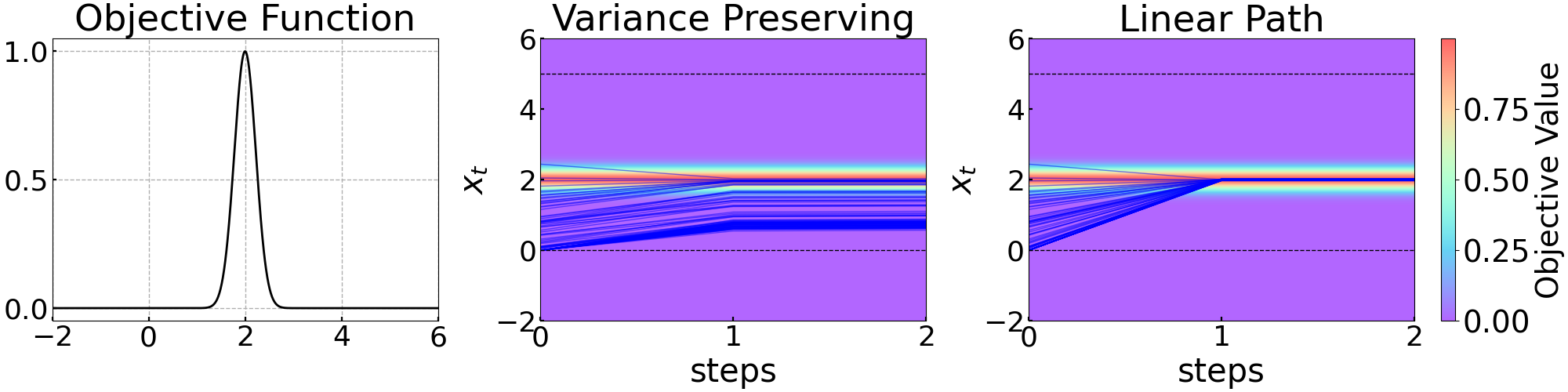}\\
    \includegraphics[scale=0.17]{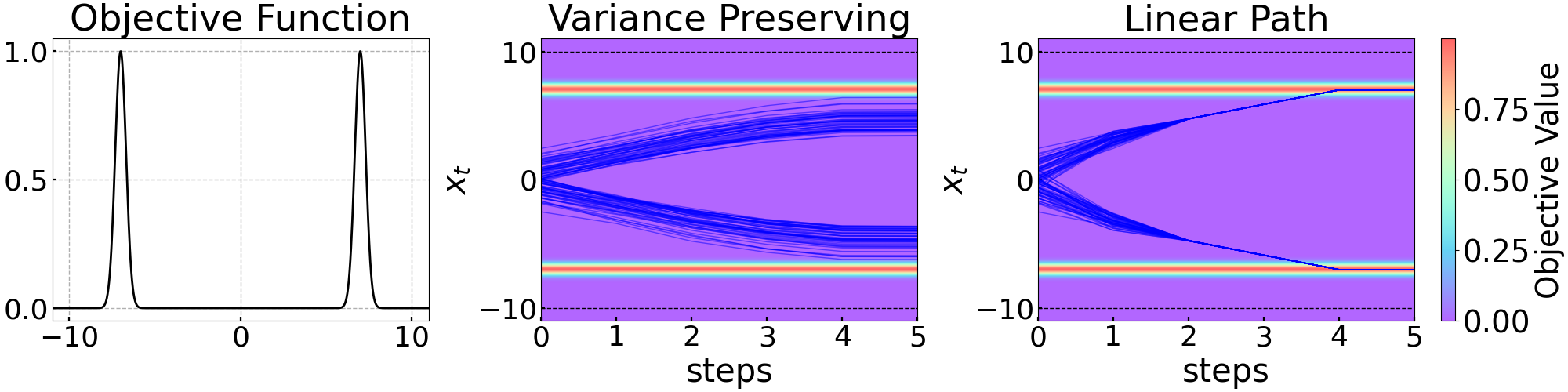}\\
    \caption{1D examples. (Top) The same example as in \cite{pan2024modelbased}. (Middle) A simple Gaussian objective function. (Bottom) A Gaussian mixture objective with two modes.}
    \label{fig:1d-numerical-examples}
\end{figure}

Fig.~\ref{fig:1d-numerical-examples} presents three one-dimensional examples. 
In the first example from \cite{pan2024modelbased}, both VP-MBD and LP-MBD reach the optimal solution within 20 steps. 
However, in the second and third Gaussian objectives, VP-MBD fails to converge, while LP-MBD rapidly reaches the optimum in only a few steps.
The performance gap between VP-MBD and LP-MBD can be attributed to differences in their noise scheduling. 
For instance, consider the second example with diffusion steps $T=2$, where we compare the standard deviations of the Gaussian proposals for VP-MBD (Eq.~\eqref{eq:mbd-gaussian-proposal}) and LP-MBD (Eq.~\eqref{eq:lp-mbd-gaussian-proposal}):
\begin{equation} \label{eq:noise-std-comparison}
\begin{split}
&\text{VP-MBD} \hspace{5mm} \sqrt{\frac{1-\bar\alpha_i}{\bar\alpha_i}} = [0.1,\,0.01] \hspace{4mm} (i=1,0),\\
&\text{LP-MBD} \hspace{5mm} \frac{t_i}{1-t_i} = [1.8,\,0.0] \hspace{4mm} (i=1,0). 
\end{split}
\end{equation}
In this case, the maximum standard deviation of LP-MBD is $\sigma_{\max}=1.8$, a value that remains unchanged regardless of the diffusion horizon $T$ since $\sigma_{\max}$ is determined independently. 
By contrast, VP-MBD with the same parameters as in the first example ($\beta_0=10^{-2}$, $\beta_1=10^{-4}$) yields a Gaussian proposal distribution with extremely small variance, thereby suppressing stochasticity and precluding meaningful exploration. 
A similar issue arises in the third example with the Gaussian mixture objective and diffusion step $T=5$, where LP-MBD again attains a maximum standard deviation of $1.8$, while VP-MBD is limited to only $0.16$.
Consequently, the change of diffusion step $T$ in VP-MBD can drastically influence its noise scale, necessitating additional tuning of $\beta_0$ and $\beta_1$ to match the behavior of LP-MBD. 
This comparison highlights that the proposed LP-MBD provides a more straightforward and intuitive parameterization, simplifying the tuning process relative to VP-MBD. Note that we only change the diffusion step $T$ and keep using the same $\beta_0$ and $\beta_1$ for VP-MBD and $\sigma_{\max}$ for LP-MBD in these three examples. 

We next evaluate ALP-MBD on a simple two-dimensional objective function.
Using REINFORCE~\cite{reinforce}, we jointly optimize the continuous noise cap $\sigma_{\max}$ and the diffusion horizon $T$, with a penalty weight $w_T=1.0$ on the step count to discourage unnecessarily large $T$.
For comparison, we apply the same reinforcement learning procedure to VP-MBD---optimizing $(\beta_0,\beta_1,T)$---and refer to this baseline as Adaptive VP-MBD (AVP-MBD).
For fair comparison, both methods are trained for the same number of policy gradient updates (30) under identical hyperparameters and evaluation budgets, and we evaluate each method using its learned parameters.

\begin{figure}[htbp]
    \centering
    \includegraphics[scale=0.22]{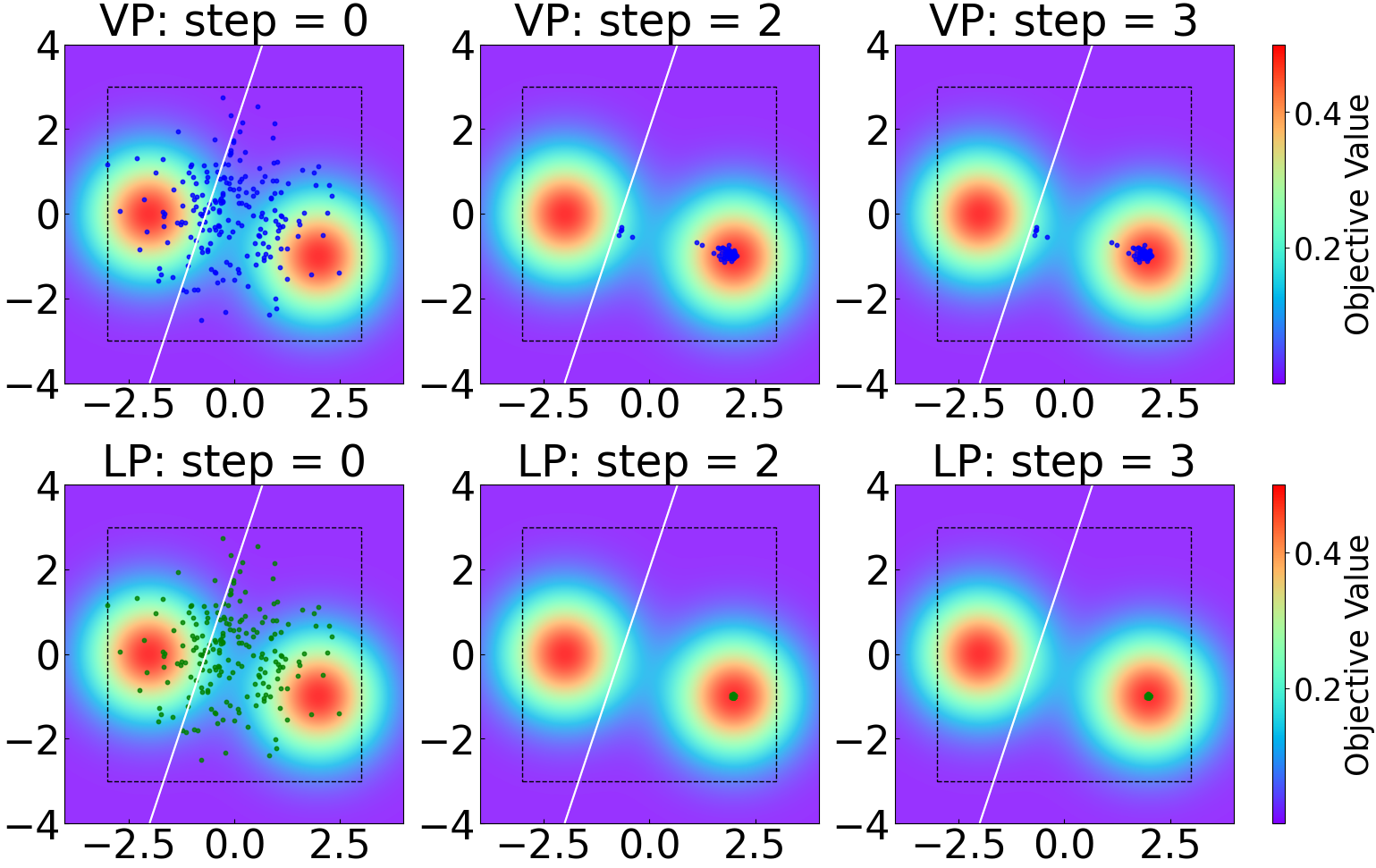}
    \caption{The comparison of AVP-MBD and ALP-MBD. The white line represents the constraint $3x_0 - x_1 \geq 2.0$. The top row shows the result of VP-MBD, and the bottom row shows the results of ALP-MBD with estimated parameters. After 30 training steps, we get $\beta_0=0.000028$, $\beta_1=0.361$, $T=3$ for AVP-MBD and $\sigma_{\max}=2.11$, $T=3$ for ALP-MBD.}
    \label{fig:2d-constrained-gaussian-mixture}
\end{figure}

Figure~\ref{fig:2d-constrained-gaussian-mixture} reports results on a 2D Gaussian mixture objective subject to a linear constraint.
At $T=3$, AVP-MBD (top) remains comparatively diffuse, with samples spread across both modes, whereas ALP-MBD (bottom) rapidly concentrates probability mass in the high-value feasible region, forming a compact cluster by step $T=3$.
These results indicate that ALP-MBD successfully discover an optimal parameter pair $(\sigma_{\max},T)$ that improves both convergence speed and solution quality, while AVP-MBD converges more slowly under the same training budget, due to the stronger parameter coupling among $(\beta_0,\beta_1,T)$ and the larger hyperparameter search space.

\subsection{Evaluation of LP-MBD}
We evaluate the performance of LP-MBD on a range of tasks in the Brax environments \cite{freeman2021brax}.
Brax is a physics-based simulator designed for large-scale reinforcement learning research, providing fast and differentiable dynamics for a variety of continuous control tasks such as locomotion, manipulation, and navigation. 

Table~\ref{tab:brax-results} summarizes the per-step rewards obtained by CEM \cite{CEM}, MPPI\cite{mppi1}\cite{mppi2}, VP-MBD\cite{pan2024modelbased}, and the proposed LP-MBD. For a fair comparison, we use the same diffusion steps $T$, horizon $H$, and sampling number. Each value represents the mean performance and its standard deviation over 5 random seeds. Overall, LP-MBD achieves competitive or superior performance across most tasks. For instance, in Hopper, HalfCheetah, and Walker2D, our approach attains the highest rewards, highlighting its effectiveness in improving control quality. 

However, it is worth noting that in the Pusher environment, LP-MBD has a lower score than VP-MBD. We hypothesize that this discrepancy arises from the higher dimensionality and increased complexity of the Pusher task. In such settings, variance-preserving noise scheduling may provide more expressive modeling capacity than a simple linear probability path, thereby yielding improved performance. We leave a more in-depth study to future work.

\begin{table}[htbp]
\caption{Per-step rewards for different tasks.}
\centering
\scriptsize
\label{tab:brax-results}
\begin{tabular}{lllll}
\toprule
  \textbf{Tasks} & \textbf{CEM} & \textbf{MPPI} & \textbf{VP-MBD} & \textbf{LP-MBD} \\
\midrule
Ant &\textbf{3.80 $\pm$ 0.43} & 2.06 $\pm$ 0.44 & 3.67 $\pm$ 0.29  & 3.76 $\pm$ 0.15 \\
Hopper & 2.24 $\pm$ 0.05 & 2.31 $\pm$ 0.05 & 2.74 $\pm$ 0.03 &  \textbf{2.74 $\pm$ 0.01}\\
HalfCheetah & 1.65 $\pm$ 2.33 & 2.10 $\pm$ 0.11 & 2.53 $\pm$ 0.31 & \textbf{2.73 $\pm$ 0.12} \\
Walker2D & 2.07 $\pm$ 0.02 & 2.08 $\pm$ 0.04 & 2.31 $\pm$ 0.02 & \textbf{2.36 $\pm$ 0.02}\\
Reacher & -0.28 $\pm$ 0.04 & -0.81 $\pm$ 0.06 & \textbf{-0.17 $\pm$ 0.04} & \textbf{-0.17 $\pm$ 0.04} \\
Pusher & -0.95 $\pm$ 0.11 & -3.47 $\pm$ 0.26 & \textbf{-0.53 $\pm$ 0.11} & -0.74 $\pm$ 0.1\\
\bottomrule
\end{tabular}
\end{table}

\subsection{Evaluation of ALP-MBD}
\begin{figure}[htbp]
    \centering
    \includegraphics[scale=0.19]{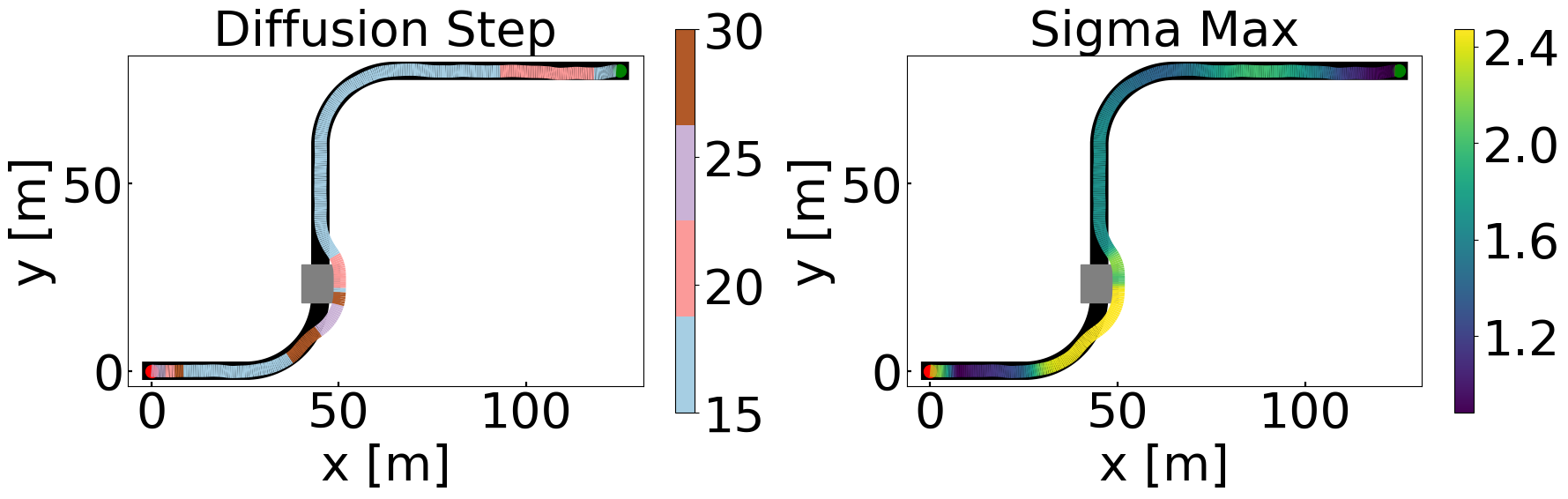}
    \caption{ALP-MBD in trajectory following tasks for a mobile robot. The red point describes the start point, and the green point indicates the goal point. The black line shows the reference trajectory, and the gray rectangle is the obstacle.}
    \label{fig:alp-mbd-evaluation}
\end{figure}

Finally, we evaluate ALP-MBD on a mobile robot trajectory-tracking task implemented in a custom gym-like environment. The robot is modeled using a kinematic vehicle model with state $x = [x, y, \theta, v]$, where $(x, y)$ denotes the position, $\theta$ is the yaw angle, and $v$ is the velocity. The control inputs are acceleration $a$ and steering rate $w$, and the dynamics evolve as
\begin{equation} \label{eq:vehicle-kinematic-equation}
\begin{split}
x_{k+1} &= x_k + v_k \cos(\theta_k) \Delta t, \\
y_{k+1} &= y_k + v_k \sin(\theta_k) \Delta t, \\
\theta_{k+1} &= \theta_k + w_k \Delta t, \\
v_{k+1} &= v_k + a_k \Delta t,
\end{split}
\end{equation}
with time step $\Delta t$. The goal is to track a predefined reference path under these dynamics while ensuring safe and smooth behavior. To quantify performance, we use a reward function that penalizes deviations from the reference trajectory, misalignment in heading, velocity error, and collisions:
\begin{equation}
\begin{split}
r = -w_{\text{lat}} \, d_{\text{lat}} &- w_{\text{yaw}} \, (\theta - \theta_{\text{ref}})^2 \\
&- w_{v} \, (v - v_{\text{ref}})^2 - w_{\text{collision}} \, \mathbf{1}_{\text{collision}},
\end{split}
\end{equation}
where $d_{\text{lat}}$ is the absolute lateral deviation from the reference path, $\psi_{\text{ref}}$ and $v_{\text{ref}}$ denote the desired yaw and velocity, $\mathbf{1}_{\text{collision}}$ is an indicator of collisions, and $w_{\bullet}$ denotes the weighting coefficients for each penalty term. The input state to the RL parameter tuning module is given by $(d_{\text{lat}}, d_{\theta}, d_{\text{vel}}, dx_{\text{obs}}, dy_{\text{obs}})$, where $d_{\theta} = \theta - \theta_{\text{ref}}$, $d_{\text{vel}} = v - v_{\text{ref}}$, and $(dx_{\text{obs}}, dy_{\text{obs}})$ denote the relative distances from the ego vehicle to surrounding obstacles. We train ALP-MBD in an environment with an S-shaped trajectory described in black line in Fig.~\ref{fig:alp-mbd-evaluation}. The gray rectangle in the figure shows a static obstacle in the environment. After the training, we compare the performance of ALP-MBD with VP-MBD and LP-MBD. For fair comparison, we set the planning horizon $H=50$ and the sampling number to 100 for all the algorithms. In this experiment, we implement each method using Python 3.11 and PyTorch.

Table~\ref{tab:alp-mbd-results} summarizes the average single-episode reward of each algorithm evaluated over five random seeds. Among the three MBD variants, the proposed adaptive method achieves the highest reward. For ALP-MBD, the reported diffusion steps $T$ and maximum noise standard deviation $\sigma_{\max}$ are averaged across the five seeds. Regarding runtime, VP-MBD and LP-MBD measure only the time to generate a control input, whereas the ALP-MBD runtime also includes the parameter estimation process (a policy forward pass to determine $T$ and $\sigma_{\max}$). This additional step accounts for its higher per-step latency relative to LP-MBD and VP-MBD. Nonetheless, the overhead introduced by parameter estimation is minor, and ALP-MBD remains well-suited for real-time control applications.

We also illustrate the trajectory generated by ALP-MBD in Fig.~\ref{fig:alp-mbd-evaluation}. ALP-MBD adaptively increases both diffusion steps and the maximum noise standard deviation when avoiding obstacles, while reducing them during steady cruising along the reference line. This adaptive behavior reflects the complexity of the underlying objective: when the ego vehicle is near an obstacle, the target objective function becomes more complex and requires additional iterations and a broader exploration range to converge to a feasible solution. In contrast, when the vehicle follows the reference trajectory in the absence of nearby obstacles, the objective remains simple, allowing convergence with fewer diffusion steps and a smaller variance. This adaptivity enhances the efficiency of VP-MBD by allocating greater computational effort only in challenging scenarios, while maintaining efficiency in simpler environments.

\begin{figure}[htbp]
    \centering
    \includegraphics[scale=0.19]{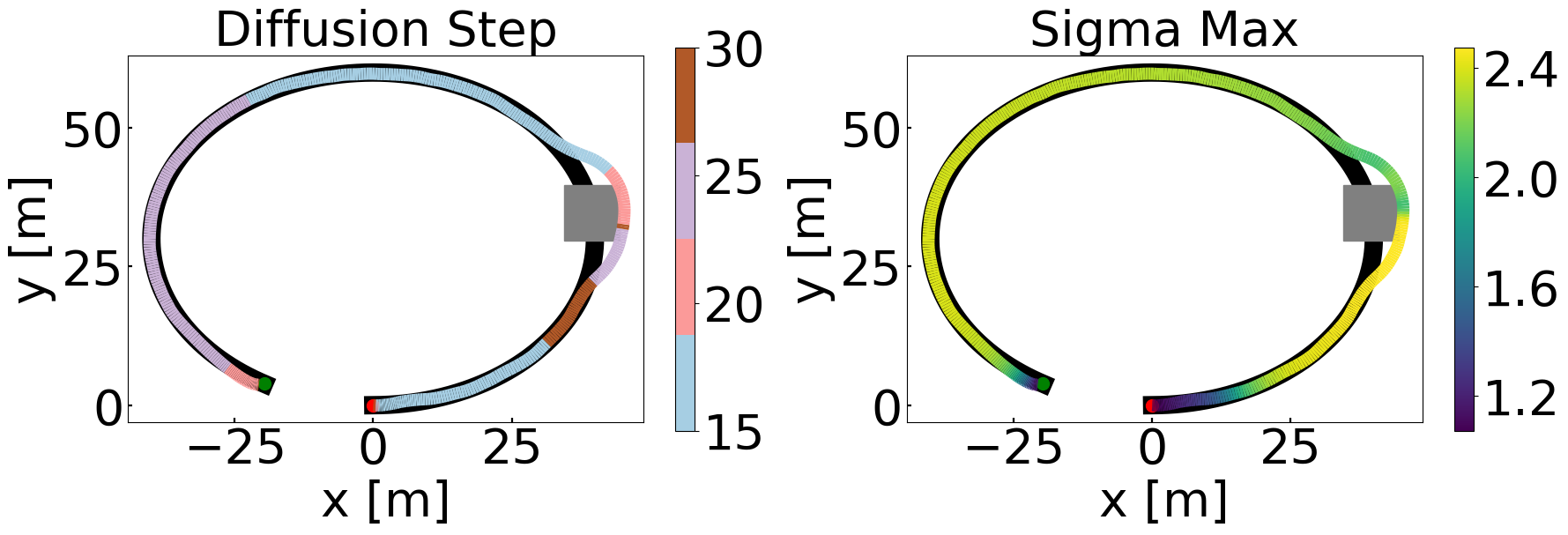}
    \caption{ALP-MBD in trajectory following tasks for a mobile robot with an oval path.  The red point describes the start point, and the green point indicates the goal point. The black line shows the reference trajectory, and the gray rectangle is the obstacle. We use the ALP-MBD trained in a different environment.}
    \label{fig:alp-mbd-generalization}
\end{figure}

In addition, we evaluate the generalization ability of the trained ALP-MBD model in a new environment with a different reference trajectory. 
As shown in Fig.~\ref{fig:alp-mbd-generalization}, ALP-MBD exhibits a consistent pattern with the previous results: it increases both the diffusion steps and the maximum noise standard deviation when the ego vehicle is near an obstacle, and decreases them when the vehicle is farther away.
We further observe that the vehicle increases diffusion steps and the maximum standard deviation when driving sharp turns in the reference trajectory. In such cases, the vehicle must apply its maximum steering angle to remain aligned with the reference path, which necessitates both additional diffusion steps and a larger noise variance. Note that ALP-MBD does not see this trajectory during training, demonstrating its ability to generalize to previously unseen scenarios.

\begin{table}[htbp]
\caption{Average single episode reward}
\centering
\label{tab:alp-mbd-results}
\scriptsize
\begin{tabular}{lllll}
\toprule
  \textbf{Methods} & Steps $T$ & $\sigma_{\max} $ & Reward  & Runtime [ms] \\
\midrule
VP-MBD & 17 & -  & -3498 $\pm$ 116  & 38.3\\
LP-MBD & 17 & 1.8 & -3465 $\pm$ 147 & \textbf{36.9} \\
ALP-MBD & 16.9 $\pm$ 1.14 & 1.82 $\pm$ 0.18 & \textbf{-3342 $\pm$ 128} & 45.5  \\
\bottomrule
\end{tabular}
\end{table}

\section{CONCLUSIONS}
We introduced Linear Path Model-Based Diffusion (LP-MBD) and its adaptive extension (ALP-MBD). LP-MBD replaces variance-preserving schedules with a flow-matching–inspired linear probability path, yielding a geometrically interpretable and decoupled parameterization. This not only reduces the burden of tuning but also provides the structural stability necessary for reinforcement learning–based adaptation. Building on this foundation, ALP-MBD dynamically adjusts diffusion steps and noise levels in response to task complexity, balancing robustness and efficiency. Through numerical examples, Brax benchmarks, and mobile robot trajectory-tracking, we demonstrated that LP-MBD simplifies scheduling while retaining strong performance, and that ALP-MBD further improves adaptability and real-time control capability.






{\small 
\bibliographystyle{IEEEtran}
\bibliography{reference}
}

\end{document}